\documentclass[10pt,journal,compsoc]{IEEEtran}
\ifCLASSOPTIONcompsoc
  \usepackage[nocompress]{cite}
\else
  \usepackage{cite}
\fi

\usepackage[latin9]{inputenc}
\usepackage{color}
\usepackage{array}
\usepackage{booktabs}
\usepackage{multirow}
\usepackage{amsmath}
\usepackage{amssymb}
\usepackage{graphicx}
\usepackage{amsmath,amssymb} 
\usepackage{color}
\usepackage{array}
\usepackage{multirow}
\usepackage{graphicx}
\usepackage{color}
\usepackage{times}
\usepackage{epsfig}
\usepackage{tabularx}
\usepackage{array}
\usepackage{algorithmic}

\usepackage[ruled,noend]{algorithm2e}
\usepackage[T1]{fontenc}
\usepackage{color}
\usepackage{booktabs}
\usepackage{amstext}
\usepackage{relsize}
\usepackage{lipsum}
\usepackage{array}
\usepackage{slashbox}

\mathchardef\mhyphen="2D 
\newcommand\fscore{\mathop{F\mhyphen score}}
\newcommand\gscore{\mathop{G\mhyphen score}}

\newcommand{\thickhline}{%
    \noalign {\ifnum 0=`}\fi \hrule height 1pt
    \futurelet \reserved@a \@xhline
}
\newcolumntype{"}{@{\hskip\tabcolsep\vrule width 1pt\hskip\tabcolsep}}
\makeatother

\newcommand\VRule[1][\arrayrulewidth]{\vrule width #1}

\newlength{\Oldarrayrulewidth}
\newcommand{\Cline}[2]{%
  \noalign{\global\setlength{\Oldarrayrulewidth}{\arrayrulewidth}}%
  \noalign{\global\setlength{\arrayrulewidth}{#1}}\cline{#2}%
  \noalign{\global\setlength{\arrayrulewidth}{\Oldarrayrulewidth}}}

\newcommand{\qed}{\nobreak \ifvmode \relax \else
      \ifdim\lastskip<1.5em \hskip-\lastskip
      \hskip1.5em plus0em minus0.5em \fi \nobreak
      \vrule height0.75em width0.5em depth0.25em\fi}

\begin{document}

\title{CGMOS: Certainty Guided Minority OverSampling}
\author{Xi Zhang, Di Ma, Lin Gan, Shanshan Jiang, Gady Agam
\thanks{X. Zhang is currently a PhD student in Illinois Institue of Technology, email: vinxi.zhang@gmail.com.}
\thanks{D. Ma, L. Gan and S. Jiang are students of Illinois Institute of Technology.}
\thanks{G. Agam is a associate professor in Illinois Institute of Technology, email: agam@iit.edu.}
}

\markboth{This paper has been acceptted by CIKM 2016.}%
{Shell \MakeLowercase{\textit{et al.}}: CGMOS: Certainty Guided Minority OverSampling}

\maketitle

\begin{abstract}
Handling imbalanced datasets is a challenging problem that if not treated correctly results in reduced classification performance. Imbalanced datasets are commonly handled using minority oversampling, whereas the SMOTE algorithm is a successful oversampling algorithm with numerous extensions. SMOTE extensions do not have a theoretical guarantee during training to work better than SMOTE and in many instances their performance is data dependent. In this paper we propose a novel extension to the SMOTE algorithm with a theoretical guarantee for improved classification performance. The proposed approach considers the classification performance of both the majority and minority classes. In the proposed approach  CGMOS (Certainty Guided Minority OverSampling) new data points are added by considering certainty changes in the dataset. The paper provides a proof that the proposed algorithm is guaranteed to work better than SMOTE for training data. Further experimental results on 30 real-world datasets show that CGMOS works better than existing algorithms when using 6 different classifiers.
\end{abstract}

\section{Introduction}
\label{sec:introduction}
In many real world problems, the distribution of data between classes is imbalanced. Learning from imbalanced datasets is an important research problem with many applications.

The fundamental issue in imbalanced learning is the ability of imbalanced data to significantly compromise the performance of standard learning algorithms \cite{HH:09}. Generally, there are three primary reasons that can cause this problem \cite{weiss2004mining}. 

The first reason is that the lack of data in the minority class makes it difficult to detect regularities within the minority class. Thus, the learned decision boundaries are less likely to approximate the true decision boundaries. 

Second, many classification algorithms utilize a general bias for better generalization and to avoid overfitting during learning. However, such bias can adversely affect the ability to learn the minority class. Inductive bias also plays a key role with respect to the minority class. Most classification algorithms prefer more common classes in the presence of uncertainty (i.e., they are biased in favor of the class priors). 

Last but not least, noise exerts a greater impact on the minority class, because in this case it is more difficult for a classifier to distinguish noise from minority data. This is especially so in extreme cases where the number of noisy samples is greater than actual minority samples. The problem of overfitting rises again, when modifying the classifier to learn the minority data correctly.

To address these problems, numerous research efforts have been devoted to imbalanced learning in recent years. The majority of techniques that solve the imbalanced learning problem fall into two categories: cost-sensitive methods and sampling-based methods. In the next section, we review related work on sampling-based methods.\footnote{Go to (https://github.com/xzhang311/CGMOS.git) for codes of this project.}

\subsection{Related work}
A number of solutions to the class-imbalance problem were previously proposed both at the data and algorithmic levels \cite{chawla2004editorial}. There are mainly three groups of methods that can solve imbalanced learning problem \cite{HH:09} including sampling methods, cost sensitive methods, and kernel methods. Sampling-based methods are very effective and easy to use when solving imbalanced learning problems. In addition, sampling-based methods can be used together with methods in the other two groups to further improve performance. In such approaches a sampling technique is used to modify an imbalanced dataset to produce a balanced distribution. It has been shown that for most imbalanced datasets, sampling techniques do improve classification accuracy.

The basic sampling methods include undersampling and oversampling. Undersampling reduces majority class samples while oversampling increases minority class samples. While several works achieving data balance through undersampling have been proposed in the past \cite{liu2009exploratory}\cite{ZJMI03}, more research efforts have been devoted to oversampling due to the fact that oversampling does not discard information.

The simplest form of oversampling is duplication of minority class samples. This approach decreases the overall level of class imbalance, but may lead to overfitting \cite{Drummond03c4.5}. SMOTE \cite{CNV:02} is a fundamental approach for oversampling using data synthesis. To balance the dataset, SMOTE randomly selects a seed sample and synthesizes a new sample by applying a linear interpolation between the seed sample and one of its neighbors. Large research efforts have been devoted to feature space data synthesis based on SMOTE. Several methods integrate data synthesis as a part of the learning procedure. For example, by introducing SMOTE in each iteration of boosting, SMOTEBoost \cite{chawla2003smoteboost} increases the number of minority class samples and focus on these cases in each boosting iteration. Using the same idea of boosting, DataBoost-IM \cite{guo2004learning} and RAMOBoost \cite{chen2010ramoboost} discover samples difficult to classify during each iteration of boosting, which are used to guide the oversampling in both the majority and minority classes. 

In another group of minority oversampling approaches, the data synthesis procedure is independent of the learning processes. Such methods give preferences to different regions of a dataset by assigning weights to samples in the dataset. These weights can then generate a probability distribution which is used for  randomly drawing samples. In such approaches the data synthesis can be completed in one step. Methods in this group include Borderline-SMOTE \cite{HH:05}, Adasyn \cite{HH:08}, \cite{barua2011novel} and MWMOTE \cite{barua2014mwmote}. All of these methods synthesize more samples along decision boundaries. However, these methods do not have objective functions to systematically guide the process of oversampling and so do not have a systematic way to decide on where new data should be synthesized. Thus, such approaches cannot measure the impact of each synthetic sample. As a result, there are several potential problems. One is that the oversampling procedure may sacrifice the performance of the majority class in order to improve the performance of the minority class in the classification. Another is that synthetic minority samples themselves can be misclassified and affect the performance in the minority class.

\subsection{Novel Contribution}
The proposed approach, CGMOS, is a member of the SMOTE family that can achieve data oversampling in a single step. To address some of the shortcomings in existing approaches, we propose a novel oversampling strategy by systematically considering the performance of both minority and majority classes. Based on a Bayesian classification framework, our proposed approach computes the influence of minority data addition on the certainty of the entire dataset. CGMOS thus can synthesize new samples that will improve the overall certainty of the entire dataset in classification. We prove that during training CGMOS is guaranteed to perform better than SMOTE when using Bayesian classification. To validate the proof, We further show experimentally that CGMOS outperforms known approaches when tested on real-world data set collections using different classifiers.

\section{Problem Formulation}
\label{sec: problem formulation}
In this paper, we address the binary classification problem for imbalanced datasets. Let $D=\{(x_j, y_j)\}_{j=1}^n$ be a training dataset, where $x_j \in \mathfrak{R}^m$ are features and $y_j \in \{l_{\mbox{mjr}}, l_{\mbox{mnr}}\}$ are ground truth class labels. We begin by formally defining the certainty of imbalanced binary classification using a Bayesian framework, where a kernel density estimation (KDE) is used to estimate the samples' probability density function (PDF). We then show how CGMOS can synthesize more samples according to the certainty estimation.

\subsection{Definition of Certainty}

Suppose $(x_j, y_j)$ is any tuple in the training dataset $D$, where $x_j$ is a feature vector and $y_j$ is the ground truth label of $x_j$.

A Bayesian classifier maps $x_j \rightarrow l$, $l\in\{l_{\mbox{mjr}}, l_{\mbox{mnr}}\}$ using following rule. 

\begin{align*}
l = \left\{
\begin{array}{cl}
  l_{\mbox{mnr}} & \text{if} \;\; \frac{P(l_{\mbox{mnr}} | x_j)}{P(l_{\mbox{mjr}} | x_j)} > 1\\
  l_{\mbox{mjr}} & \text{otherwise}
\end{array}\right.
\end{align*}
where the posterior probability $P(l|x_j)$ is computed using Bayes' rule:
\begin{align*}
P(l | x_j) = \frac{P(x_j | l)P(l)}{P(x_j)}; \;\;\;\; l\in\{l_{\mbox{mjr}}, l_{\mbox{mnr}}\}
\end{align*}
 
Uncertainty is commonly used in machine learning algorithms. In this work, we use the posterior probability $P(y_j | x_j)$ to define certainty. This is because in classification, the posterior probabilities $P(y_j | x_j)$ reflect the certainty of assigning a sample to a correct label, where higher numbers indicate classification results with a stronger certainty.
\newline
\newline
\noindent \textbf{Definition 1. (Certainty)} Let $(x_j, y_j)$ be any tuple in $D$, where $x_j$ is a feature vector and $y_j$ is the ground truth label of $x_j$. The certainties for samples in the majority and minority class are respectively defined as:

\begin{equation}
C(y_j=l_{\mbox{mjr}} | x_j) = P(y_j=l_{\mbox{mjr}}|x_j)
\end{equation}

\begin{equation}
C(y_j=l_{\mbox{mnr}} | x_j) = P(y_j=l_{\mbox{mnr}}|x_j)
\end{equation} 

It should be noted that in the case of binary classification the definition of certainty above is related up to some constants to the uncertainty defined in \cite{2013Sharma} based on margin confidence.

\subsection{PDF Estimation}
There are two general ways to estimate a density function: parametric or non-parametric. In this work we use a non-parametric model so as to not depend on a specific distribution model. We use kernel density estimation (KDE)\cite{elgammal2002background}\cite{zhang2006bayesian} to estimate the likelihood $P(x_j|l)$, $l\in \{l_{\mbox{mjr}}, l_{\mbox{mnr}}\}$.

Assuming that the data is independent and identically distributed (i.i.d) and drawn from some distribution with an unknown density $P(x_j | l)$, we have using KDE:

\begin{equation}
\begin{split}
P(x_j | l) =& \frac{\sum_{k=1}^n K(\frac{x_j-x_k}{h_k}) \mathrm{I}(y_k=l)}{\sum_{k=1}^n \mathrm{I}(y_k=l)}
\end{split}
\end{equation}
where $l \in \{l_{\mbox{mjr}}, l_{\mbox{mnr}}\}$, $I(\cdot)$ is an indicator function, and $K(\cdot)$ is a kernel function which has zero mean and integrates to one. Given any sample $x_k$, the bandwidth $h_k$ of the sample $x_k$ controls the effective range of the kernel and smoothness of the density function. Intuitively one wants to choose $h_k$ as small as the data allows to exhibit as many underlying structures of the data as possible. Small bandwidth, however, will result in a noisy estimate. In this work, for any sample $x_k$, we calculate a bandwidth $h_k$ as a scaled average distance between $x_k$ and its $q$ nearest neighbors:

\begin{equation}
h_k = \sigma \cdot \frac{\sum_{x \in N(x_k)} \| x-x_k \|}{q}
\end{equation}
where $N(x)$ is the set of the $q$ nearest neighbors of $x_k$ and $\sigma > 0$ is a scale factor applied to the distance. We will discuss selection of parameters $\sigma$ and $q$ in Section \ref{sec: results}.

\subsection{Oversampling Seed Selection}
In most classification algorithms, samples close to decision boundaries have less certain classification results. In order to achieve better predictions for such samples, many existing approaches synthesize data directly along the boundaries. However, this is risky and the expected performance improvement is not guaranteed. There are two primary reasons. First, samples from both classes are mixed in regions near the boundaries. Synthetic samples if added to  these regions are less predictable and hard to learn. Second, adding synthetic minority samples to these regions may adversely impact the majority class, which may in turn decrease the performance of the majority class in classification. Instead of unguided oversampling near the boundaries, our proposed approach targets adding samples by considering the certainties of both the minority and majority classes before and after adding the samples. The synthetic samples thus are added to locations that can improve the overall certainty of the original data and boost the performance of the classification.

CGMOS uses a similar procedure as SMOTE when synthesizing a new sample. The sample is produced by interpolating between one seed sample and some of its neighbors. However, instead of randomly drawing a seed sample for interpolation, CGMOS assigns each sample $(x_i, y_i) \in D$ a weight $W(x_i)$ which is used to determine the probabilities of $x_i$ being chosen for interpolation. A higher weight results in a higher probability of a point being selected.

To compute $W(x_i)$, we suppose that a new sample will be added to the same location as $x_i$. The weight $W(x_i)$ is computed as a relative certainty change\footnote{Measuring absolute certainty increments will not work, because measuring magnitude will give higher preference to parts which already have high certainty.} comparing the certainty before and after the sample is added. With a new sample added at location $x_i$, we update the certainty for all $(x_j, y_j) \in D$ and denote it as $C_{+i}(y_j | x_j)$. 
\newline
\newline
\noindent \textbf{Definition 2. (Relative Certainty Change)} The relative certainty change of label $y_j$ assigned to feature $x_j$ due to adding a minority example at location $x_i$ is defined by:

\begin{equation}
R_{+i}(y_j | x_j)= \frac{C_{+i}(y_j | x_j) -C(y_j | x_j)}{C(y_j | x_j)}
\label{eqn: relative diff}
\end{equation}
where $C(y_j | x_j)$ is the certainty before addition.

When computing $W(x_i)$, CGMOS considers the relative certainty changes of examples from both the majority and the minority classes. $W(x_i)$ is computed as the average value of relative certainty changes of all samples in the dataset.

\begin{equation}
\begin{split}
W(x_i)= 1+\frac{1}{n}\sum_{j=1}^n R_{+i}(y_j | x_j)
\end{split} 
\label{eqn: weight}
\end{equation}

Given $W(x_i)$ for all $x_i \in D$, it is easy to see $W(x_i)>0$. We compute a normalization factor $z$ so that $\frac{1}{z}\sum_{i=1}^n W(x_i)=1$. Therefore, the oversampling procedure can randomly choose sample for interpolation according $W(x_i)/z$. The interpolation phase of CGMOS is  the same as SMOTE\cite{CNV:02}. 

\begin{figure}[ht]
\centerline{ %
\begin{tabular}{c}
\resizebox{0.48\textwidth}{!}{\rotatebox{0}{ 
\includegraphics{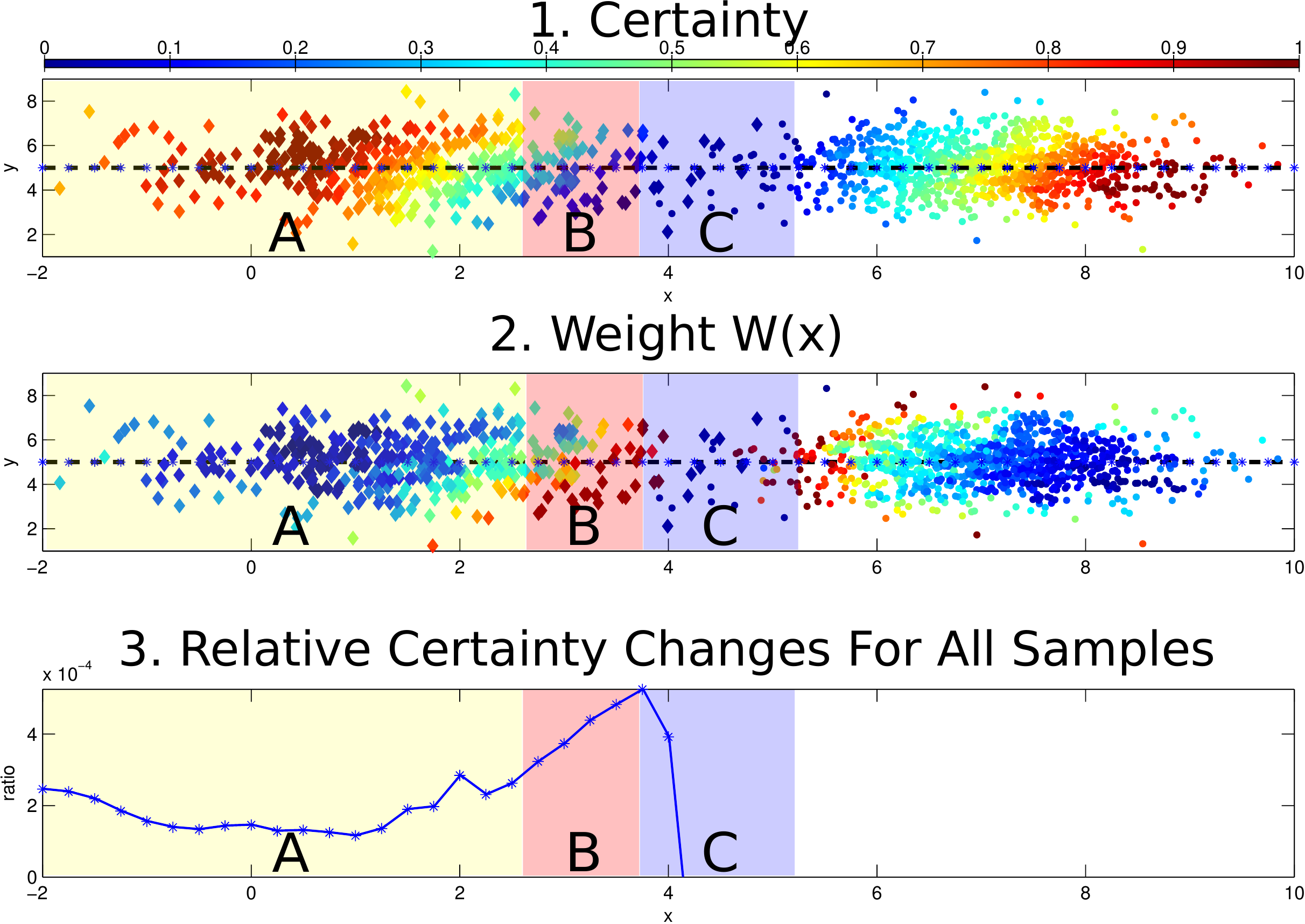}}}
\end{tabular}} 
\caption{Demonstration of CGMOS. In first two figures, diamonds represent minority samples and circles represent majority samples. The positions of synthesized data points are labeled using a star symbol on a horizontal line passing through the center. The x and y axes represent features. In the bottom figure the x axis indicates a location where a sample was added (in correspondence with the first two figures) whereas the y-axis indicates the relative certainty change.}
\label{fig: demoofdiffinsertion} 
\end{figure}

A demonstration of CGMOS is shown in Fig. \ref{fig: demoofdiffinsertion}. In this figure, samples in both the majority and minority classes are randomly drawn based on Gaussian distribution, where the means of the two datasets are on the same horizontal line, and the mean of the majority is to the right of the minority. The majority class contains 2000 samples and the minority class contains 400 samples. Color in part 1 of the figure indicates the certainty of each example with respect to its class, where red indicates high certainty. We highlight 3 regions (A, B, C) in the minority class. Samples in region A have relative high certainties, sample in region B has low certainties and region C is a boundary region in which samples have the lowest certainties. Part 2 of the figure shows the weight of each example as computed by our approach where red indicates high values. Region B has higher values and is where CGMOS will synthesize most of the samples.

To show the certainty changes induced by adding samples at different locations of the dataset, in part 3 of the figure we add one minority sample and move its location with a fixed step size from left to right on a horizontal line passing through the two classes. We then compute the relative certainty changes for all samples in both classes. As can be observed, by measuring relative certainty changes, CGMOS will assign a higher weight to samples in region B. The figure also shows that by oversampling more in region B, the certainty of the entire dataset gets improved, because the relative certainty changes are positive.

\section{Theoretical Guarantee Over SMOTE}
Several existing approaches claim handling imbalanced learning better than SMOTE. Such claims are normally validated using empirical tests without a theoretical guarantee and in some instances may not extend to new datasets. In this section we provide a theoretical guarantee showing that CGMOS is expected to work better than SMOTE in training process.

Let $D=\{(x_j, y_j)\}_{j=1}^n$ be a training dataset. Let $W(D)=\{W(x_i)\}_{i=1}^n$ be the sample weights computed using Eqn. \ref{eqn: weight}. 
\newline
\newline
\textbf{Lemma 1.} \textit{Given a set of weights $\{W(x_i)\}_{i=1}^n$ as defined above and a normalization factor $z$ given by $z=\sum_{i=1}^n W(x_i)$, it must be that $\sum_{i=1}^n W(x_i)^2 \geq \frac{z^2}{n}$.}
\newline
\newline
\textbf{Proof}
Let $W$ be an n-dimensional vector whose elements are $W(x_i)$. Let $I$ be an n-dimensional vector whose elements are all 1. Using the Cauchy-Schwarz inequality we have: $|W\cdot I|\leq \|W\|\cdot\|I\|$. Thus, $|\sum_{i=1}^n W(x_i)|\leq \sqrt{\sum_{i=1}^n W(x_i)^2 }\sqrt{n}$ using the fact that $\sum_{i=1}^n W(x_i)=z$, we thus have $\sum_{i=1}^n W(x_i)^2 \geq \frac{z^2}{n}$.  $\;\;\;\blacksquare$
\newline

\noindent\textbf{Definition 3. (Addition Likelihood Ratio)} Let $\theta$ denote the non-parametric likelihood estimate $P(x_j|l)$, $l\in \{l_{\mbox{mjr}}, l_{\mbox{mnr}}\}$ before a new sample $x_i$ is added, and $\theta'$ denote the non-parametric likelihood estimate after the new sample is added. The addition likelihood ratio $r_{+i}(y_j|x_j)$ of example $x_j$ by adding data to $x_i$ location is defined as the ratio between the likelihood estimate after the new addition and the likelihood estimate before the new addition: 
\begin{equation}
r_{+i}(y_j|x_j) \equiv P(y_j|x_j ; \theta') / P(y_j|x_j; \theta).
\label{eqn. likelihoodratio}
\end{equation}

\noindent \textbf{Lemma 2.} \textit{The addition likelihood ratio $r_{+i}(y_j|x_j)$ is related to the relative certainty change ratio $R_{+i}(y_j|x_j)$ by: }
\begin{equation}
r_{+i}(y_j|x_j)=1+R_{+i}(y_j|x_j).
\end{equation}

\noindent\textbf{Proof} According to the definition of the certainty, we have $C_{+i}(y_j|x_j)=P(y_j|x_j; \theta')$ and $C(y_j|x_j; \theta)=P(y_j|x_j; \theta)$. Then $P(y_j|x_j; \theta')=r_{+i}(y_j|x_j) P(y_j|x_j; \theta)$ according to the definition of likelihood ratio. Given Eqn. \ref{eqn: relative diff}, we have that $R_{+i}(y_j | x_j)=\frac{r_{+i}(y_j|x_j) P(y_j | x_j; \theta) - P(y_j |x_j; \theta)}{P(y_j |x_j; \theta)}$.  By simplifying this equation, we thus have 
\begin{equation}
r_{+i}(y_j|x_j)=1+R_{+i}(y_j|x_j). \;\;\;\blacksquare
\end{equation}
The addition likelihood ratio defined in Eqn. \ref{eqn. likelihoodratio} measures the gain in adding a new point, where higher gains are desired. Note that while the gain is normally close to 1 it may be bigger or smaller than 1.
\newline

\noindent\textbf{Definition 4. (Average gain)} The average gain when adding sample $x_i$ is defined by:
\begin{equation}
\bar{r}_{+i}=\frac{1}{n} \sum_{j=1}^n r_{+i}(y_j|x_j)
\end{equation}

\noindent\textbf{Lemma 3.} \textit{Given the average gain, it must be that:}
\begin{equation}
\bar{r}_{+i}=W(x_i).
\end{equation}
\noindent\textbf{Proof} Using the definition of $W(x_i)$ we have $W(x_i)=\frac{1}{n}\sum_{j=1}^n R_{+i}(y_j|x_j)$. Using Lemma 2 we can replace $r_{+i}(y_j | x_j)-1$ with $R_{+i}(y_j | x_j)$. Hence:
\begin{equation}
\bar{r}_{+i}=\frac{1}{n}\sum_{j=1}^n R_{+i}(y_j | x_j) + 1\equiv W(x_i) \;\;\;\blacksquare
\end{equation}

The average gain is an indicator of the benefit of CGMOS. We show that the expected average gain is higher in proposed approach compared with SMOTE.
\newline

\noindent \textbf{Theorem 1.} \textit{The expected average gain in CGMOS is higher or equal to that of SMOTE.}
\newline

\noindent\textbf{Proof} For CGMOS the expected average gain is given by:
\begin{equation}
E_p\equiv E[\bar{r}_{+i}] = \sum_{i=1}^n \bar{r}_{+i} \frac{W(x_i)}{z}
\end{equation}
where $z$ is the normalization factor as defined earlier. 
Using Lemma 3:
\begin{equation}
E_p=\sum_{i=1}^n W(x_i)\frac{W(x_i)}{z}=\frac{1}{z}\sum_{i=1}^n W^2(x_i).
\end{equation}
\newline
\noindent For SMOTE the expected average gain is given by:
\begin{equation}
E_s\equiv E[\bar{r}_{+i}] = \sum_{i=1}^n \bar{r}_{+i} \frac{1}{n}
\end{equation}
Using Lemma 3:
\begin{equation}
E_s = \frac{1}{n}\sum_{i=1}^n W(x_i) =\frac{z}{n}
\end{equation}
Using Lemma 1:
\begin{equation}
E_p= \frac{1}{z}\sum_{i=1}^n W^2(x_i)\geq \frac{1}{z}\frac{z^2}{n}= E_s \;\;\;\blacksquare
\end{equation}

\section{Results and Discussion}
\label{sec: results}

\subsection{Datasets}
30 real-world datasets were randomly chosen from the UCI machine learning repository \cite{Lichman:2013} for empirical testing of CGMOS. Most of the datasets were released within the past 10 years. As some of the datasets contain samples of more than two classes, we convert such datasets to a binary classification problem by keeping the class with the least data and merging all other classes. A summary of the test collections is provided in Table \ref{tab: realdata}.

\begin{table*}[ht]
\begin{center}
\scalebox{1}
{
\begin{tabular}{!{\VRule[1.5pt]}l!{\VRule[1.5pt]}l|l|l|l!{\VRule[1.5pt]}l!{\VRule[1.5pt]}l|l|l|l!{\VRule[1.5pt]}}
\specialrule{1.5pt}{0pt}{0pt} 
\multicolumn{1}{!{\VRule[1.5pt]}c!{\VRule[1.5pt]}}{{\bf Name}} & \multicolumn{1}{c|}{{\bf S \#}} & \multicolumn{1}{c|}{{\bf F \#}} & \multicolumn{1}{c|}{{\bf R}} & \multicolumn{1}{c!{\VRule[1.5pt]}}{{\bf Year}} & \multicolumn{1}{c!{\VRule[1.5pt]}}{{\bf Name}} & \multicolumn{1}{c|}{{\bf S \#}} & \multicolumn{1}{c|}{{\bf F \#}} & \multicolumn{1}{c|}{{\bf R}} & \multicolumn{1}{c!{\VRule[1.5pt]}}{{\bf Year}}\\ 
\specialrule{1.5pt}{0pt}{0pt} 
\textbf{BankMarket} & 45211 & 17 & 0.13 & 2012 & \textbf{Libras} & 360 & 91 & 0.07 & 2009 \\ \hline
\textbf{BloodService} & 748 & 5 & 0.31 & 2008 & \textbf{MultipleFs} & 2000 & 649 & 0.11 & 1998 \\ \hline
\textbf{BreastCancer} & 400 & 9 & 0.53 & 1988 & \textbf{Parkinson} & 1040 & 26 & 0.02 & 2014 \\ \hline
\textbf{BreastTissue} & 106 & 10 & 0.15 & 2010 & \textbf{PlanRelax} & 182 & 13 & 0.4 & 2012 \\ \hline
\textbf{CarEvaluation} & 1730 & 6 & 0.04 & 1997 & \textbf{QSAR} & 1055 & 41 & 0.51 & 2013 \\ \hline
\textbf{Card'graphy} & 2126 & 23 & 0.09 & 2010 & \textbf{SPECT} & 268 & 22 & 0.26 & 2001 \\ \hline
\textbf{CharacterTraj} & 2860 & 3 & 0.04 & 2008 & \textbf{SPECTF} & 134 & 44 & 0.26 & 2001 \\ \hline
\textbf{Chess} & 3198 & 22 & 0.91 & 1989 & \textbf{SeismicBumps} & 2584 & 19 & 0.07 & 2013 \\ \hline
\textbf{ClimateSim} & 540 & 18 & 0.09 & 2013 & \textbf{Statlog} & 2310 & 19 & 0.17 & 1990 \\ \hline
\textbf{Contraceptive} & 1474 & 9 & 0.29 & 1997 & \textbf{PlatesFaults} & 1941 & 27 & 0.03 & 2010 \\ \hline
\textbf{Fertility} & 100 & 10 & 0.14 & 2013 & \textbf{TAEvaluation} & 151 & 5 & 0.49 & 1997 \\ \hline
\textbf{Haberman} & 306 & 3 & 0.36 & 1999 & \textbf{UKnowledge} & 403 & 5 & 0.1 & 2013 \\ \hline
\textbf{ILPD} & 580 & 10 & 0.4 & 2012 & \textbf{Vertebral} & 310 & 6 & 0.48 & 2011 \\ \hline
\textbf{ImgSeg} & 2310 & 19 & 0.17 & 1990 & \textbf{Customers} & 440 & 8 & 0.48 & 2014 \\ \hline
\textbf{Leaf} & 342 & 16 & 0.24 & 2014 & \textbf{Yeast} & 1484 & 8 & 0.04 & 1996 \\ \specialrule{1.5pt}{0pt}{0pt} 
\end{tabular}
}
\end{center}
\caption{Summary of the datasets used in our experiments, where S\#, F\#, and R stand for the number of samples, the number of features, and imbalance ratio (defined as \#minority/\#majority).}
\label{tab: realdata}
\end{table*}

\begin{figure}[ht]
\centerline{ %
\begin{tabular}{cc}
\resizebox{0.23\textwidth}{!}{\rotatebox{0}{ 
\includegraphics{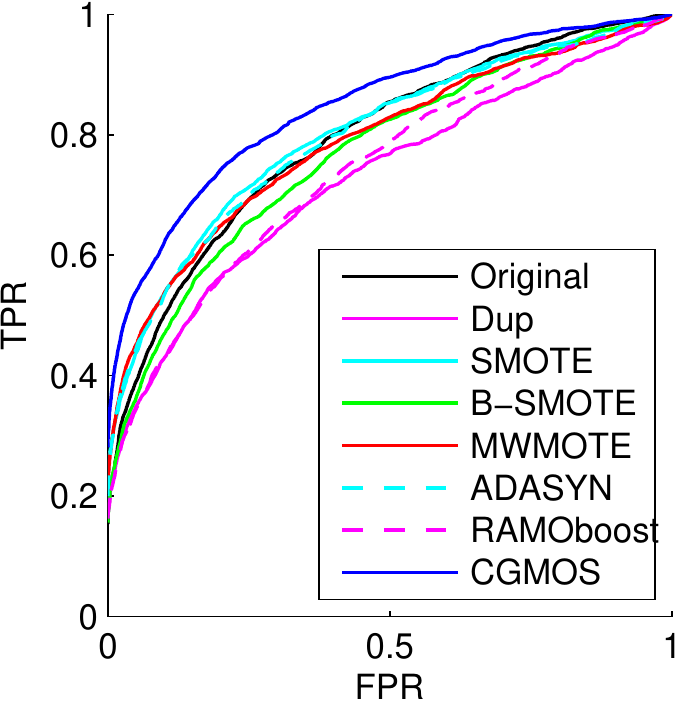}}}
&
\resizebox{0.23\textwidth}{!}{\rotatebox{0}{ 
\includegraphics{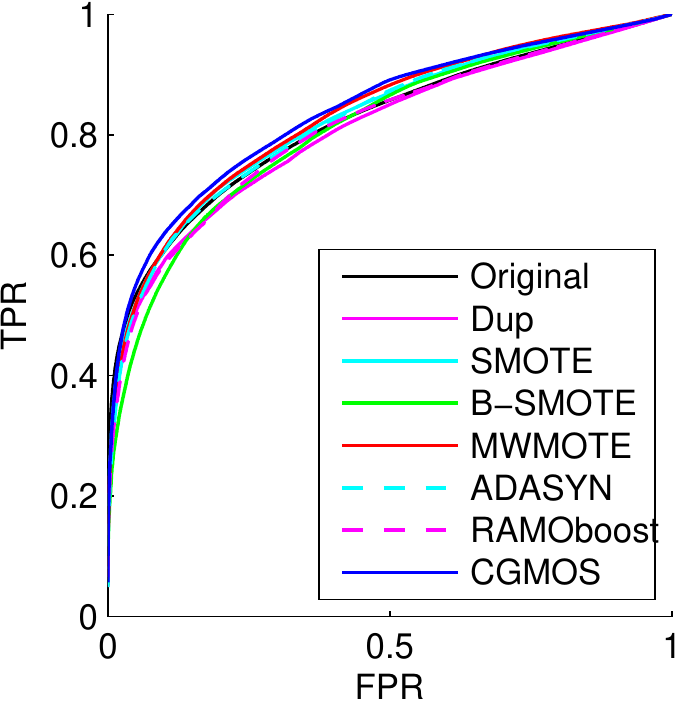}}}
\\
b-kde & knn
\\
\resizebox{0.23\textwidth}{!}{\rotatebox{0}{ 
\includegraphics{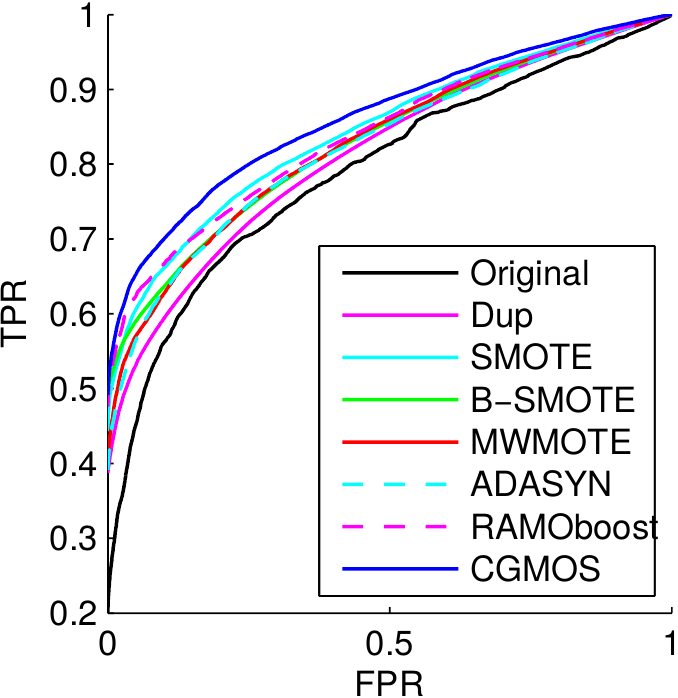}}}
&
\resizebox{0.23\textwidth}{!}{\rotatebox{0}{ 
\includegraphics{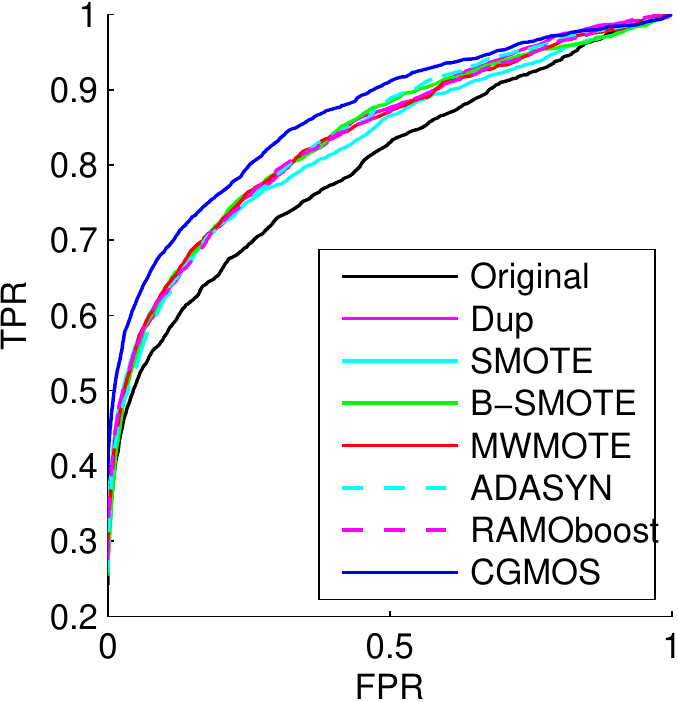}}}
\\
svm & nn
\\
\resizebox{0.23\textwidth}{!}{\rotatebox{0}{ 
\includegraphics{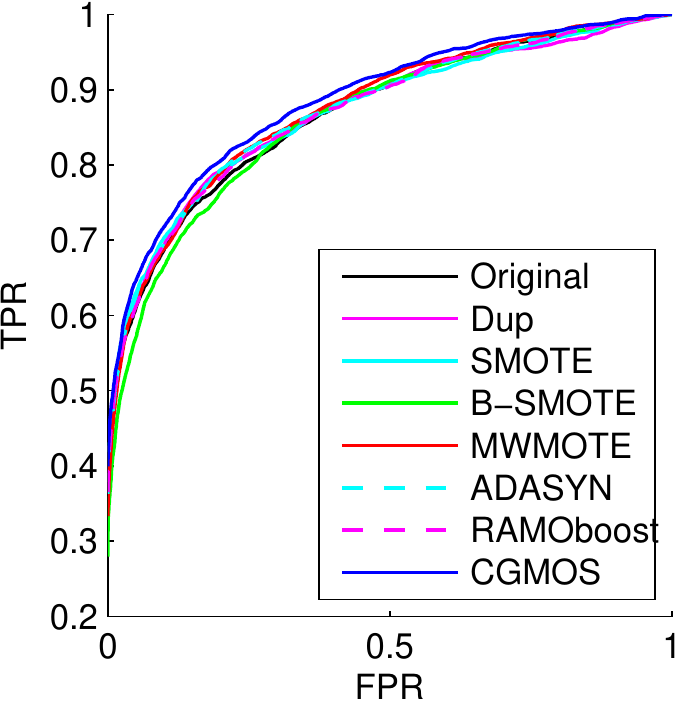}}}
&
\resizebox{0.23\textwidth}{!}{\rotatebox{0}{ 
\includegraphics{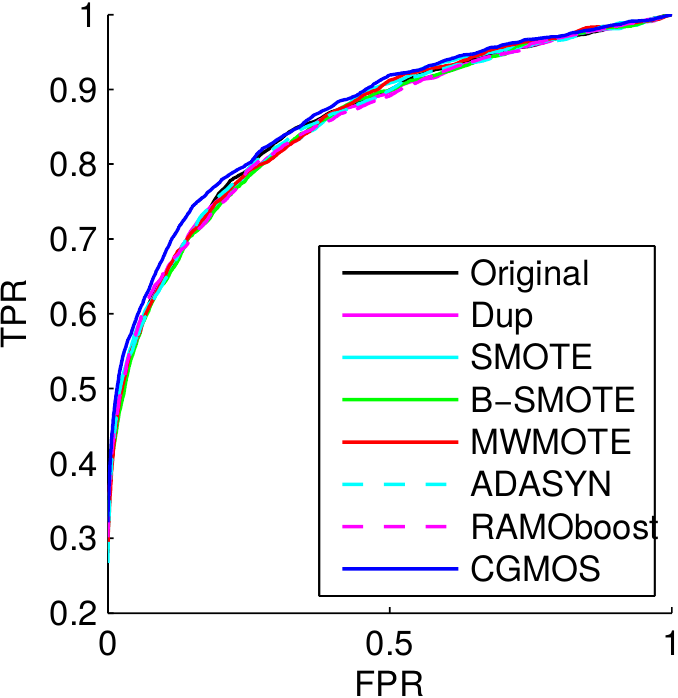}}}
\\
rf & Adaboost.M1
\\
\end{tabular}} 
\caption{ROC curves of classification results. From left to right, up to down, we show the results of 6 different classifiers: b-kde, knn, svm, nn, rf and Adaboost.M1. Curves in blue are the results of the proposed CGMOS.}
\label{fig: roc} 
\end{figure}

\begin{figure}[t]
\centerline{ %
\begin{tabular}{ccc}
\resizebox{0.23\textwidth}{!}{\rotatebox{0}{ 
\includegraphics{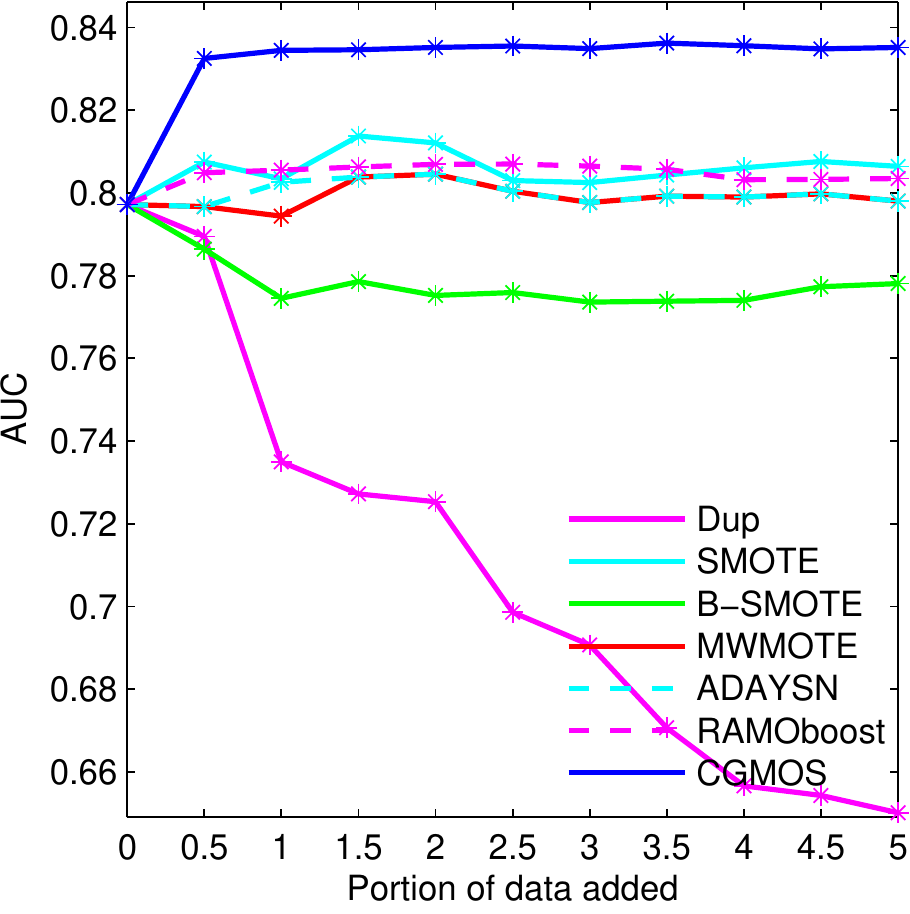}}}
&
\resizebox{0.23\textwidth}{!}{\rotatebox{0}{ 
\includegraphics{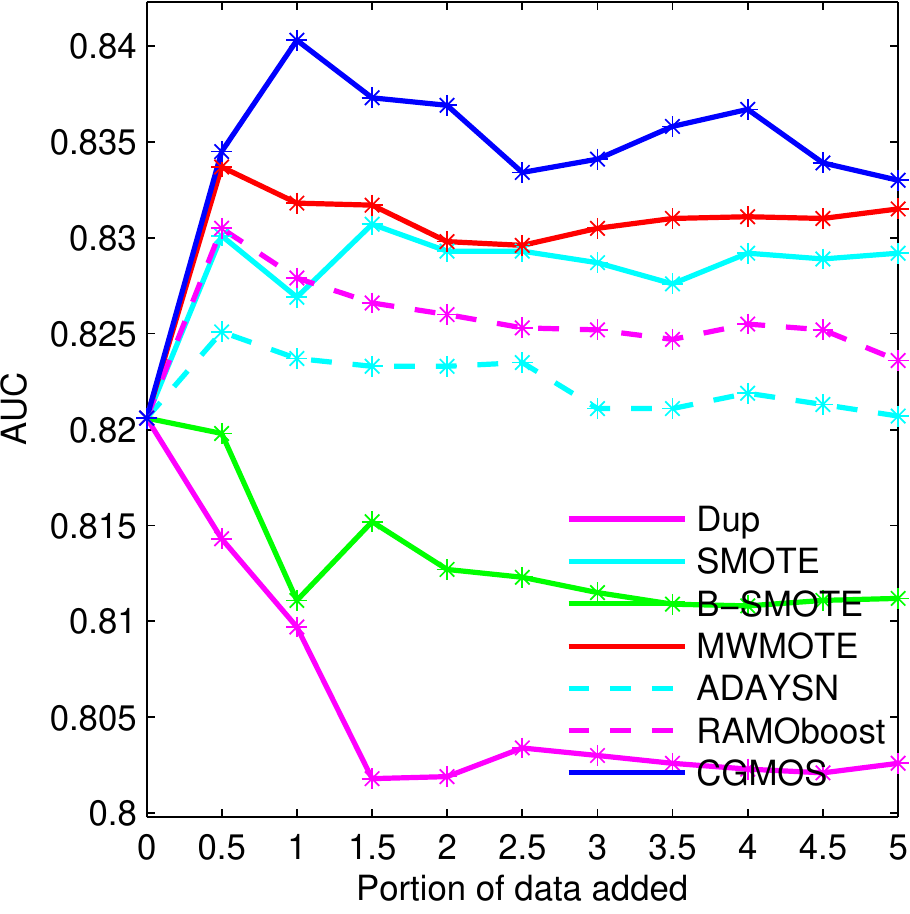}}}
\\
b-kde & knn
\\
\resizebox{0.23\textwidth}{!}{\rotatebox{0}{ 
\includegraphics{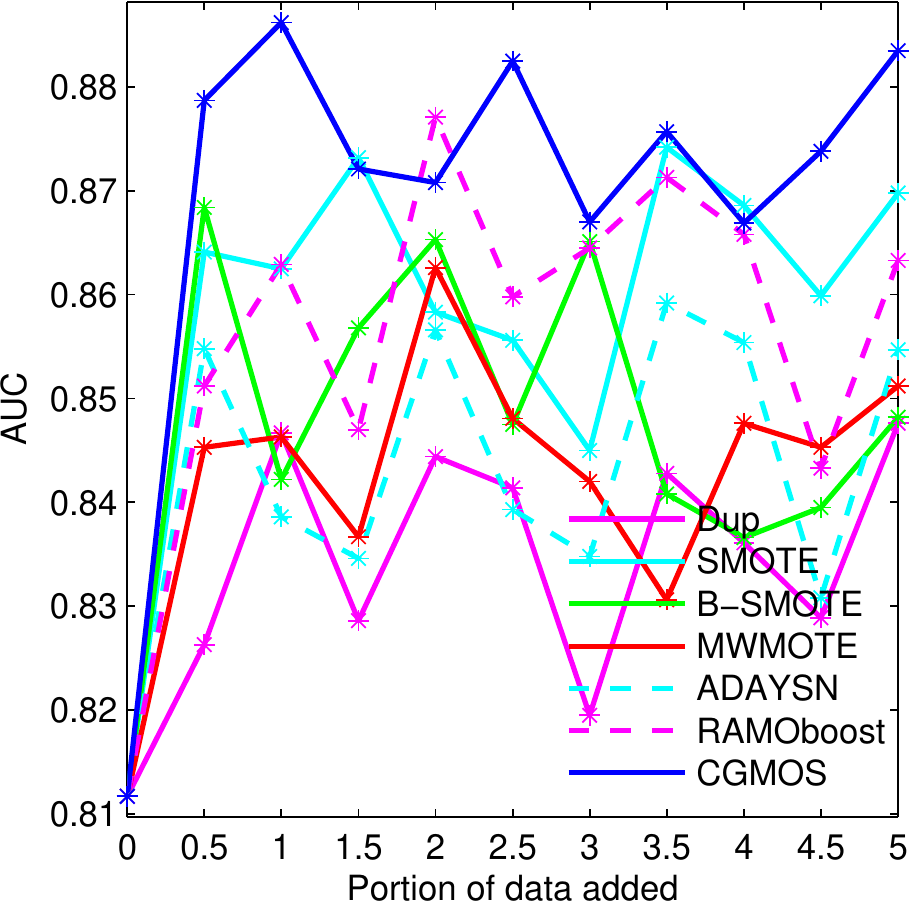}}}
&
\resizebox{0.23\textwidth}{!}{\rotatebox{0}{ 
\includegraphics{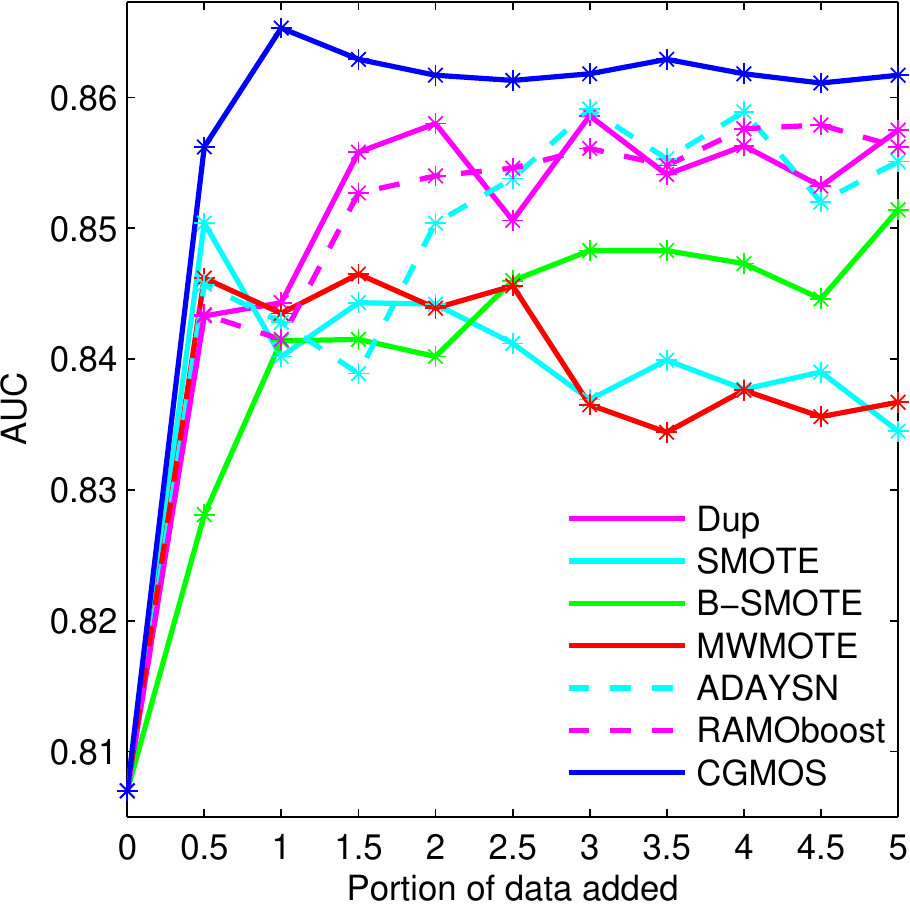}}}
\\
svm & nn
\\
\resizebox{0.23\textwidth}{!}{\rotatebox{0}{ 
\includegraphics{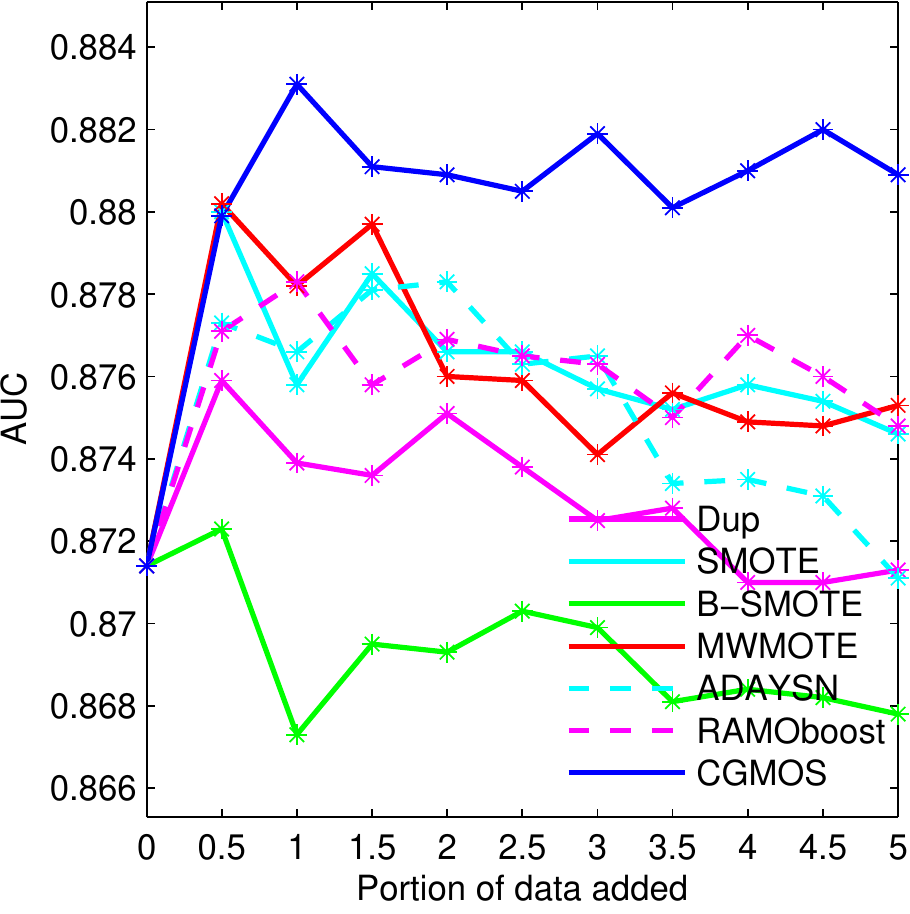}}}
&
\resizebox{0.23\textwidth}{!}{\rotatebox{0}{ 
\includegraphics{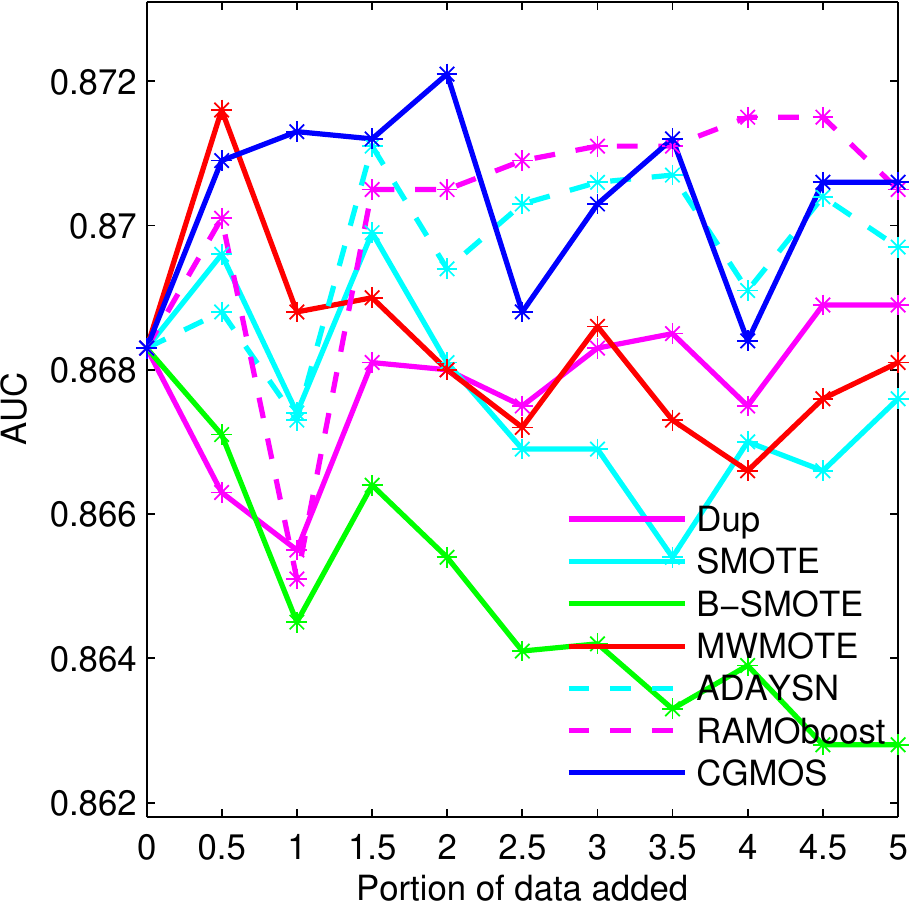}}}
\\
rf & Adaboost.M1
\\
\end{tabular}} 
\caption{Comparison of results when increasing the number of data synthesized for the minority class. The curves measure the average AUC of the ROC curves. Curves in blue are the results of the proposed CGMOS.}
\label{fig: increasingnum} 
\end{figure}

\subsection{Compared Approaches}
According to a survey of imbalanced learning \cite{HH:09}, there are mainly three groups of methods addressing imbalanced learning: sampling methods, cost sensitive methods, and kernel methods. The proposed CGMOS belongs to the sampling group. Thus, we compare CGMOS to five other oversampling methods in this group: SMOTE\cite{CNV:02}, Borderline-SMOTE\cite{HH:05}, ADASYN\cite{HH:08}, MWMOTE\cite{barua2014mwmote} and RAMOBoost\cite{chen2010ramoboost}. Since oversampling by duplication is broadly used in many applications as a baseline, we add it to our evaluation as well. To demonstrate the improvement of these oversampling strategies, we include in the comparison raw data with no oversampling. It should be noted that sampling methods are often combined with cost sensitive methods and kernel methods to further boost learning. \cite{chawla2004editorial}\cite{chawla2003smoteboost}\cite{guo2004learning}.

\subsection{Base classifiers}
We match the compared classifiers to classifiers used in other SMOTE extension evaluations. Six well-known classifiers are tested in experiments. The first is the Bayesian classifier based on kernel density estimation described in Section \ref{sec: problem formulation} (b-kde). The second is a K nearest neighbors classifier (knn). The third is a support vector machine classifier using RBF kernel (svm). The fourth one is a neural network (nn) with one hidden layer. We use in addition two ensemble methods: a random forest implementing the C4.5 decision tree \cite{Quinlan:1993} (rf) and Adaboost.M1 \cite{fy:1996}. All hyper-parameters of the classifiers tested were determined by cross validation to ensure the best performance of each method.

\subsection{Evaluation metric}
Finding an appropriate evaluation metric for different tasks is challenging, since different evaluation metrics are designed for different purposes. The datasets used in this paper cover from financial application to medical treatment. To achieve an general evaluation and avoid bias, we follow the method in \cite{CNV:02}\cite{HH:05}\cite{HH:08}\cite{barua2014mwmote}\cite{chen2010ramoboost} and use different metrics to evaluate the performance of the proposed CGMOS oversampling algorithm.

Among these evaluation metrics, the most frequently adopted ones are $Precision$ and $Recall$ when the focus of evaluation is focus on one specific class such as problems in text classification, information extraction, natural language processing and bioinformatics. In these areas of application the number of examples belonging to one class is often substantially lower than the overall number of examples, which basically are imbalance learning problems. $Precision$ and $Recall$ are defined as:

\begin{align*}
Precision &= \frac{TP}{(TP + FP)}\\
Recall &= \frac{TP}{(TP+FN)}  
\end{align*}

However, these two metrics share an inverse relationship between each other. A quick inspection on the $Precision$ and $Recall$ formulas readily yields that solely use each of these two metrics only provide a limit view of an algorithm under test. As $Recall$ provides no insight to how many examples are incorrectly labeled as positive and $Precision$ cannot assert how many positive examples are labeled incorrectly. Specifically, the $\fscore$ combines $Precision$ and $Recall$ as measure of the effectiveness of classification in terms of a ration of the weighted importance on either $Recall$ or $Precision$, which is defined as:
\begin{align*}
\fscore = \frac{(1+\beta^2)\cdot Precision \cdot Recall}{(\beta^2\cdot Precision)+ Recall}.
\end{align*}
We use $\beta=1$ to treat $Precision$ and $Recall$ equally in all evaluations. As a result, $\fscore$ provides more insight into the functionality of a classifier.

As $\fscore$ measures the harmonic mean of $Precision$ and $Recall$, we also compute $Gscore$ which is the geometric mean of $Precision$ and $Recall$ and is able to evaluate the degree of inductive bias in terms of a ratio of positive accuracy and negative accuracy \cite{HH:09}.

\begin{align*}
\gscore=\sqrt{Precision \cdot Recall}
\end{align*}

As both $\fscore$ and $\gscore$ concentrate their measures on one class (positive examples) \cite{sokolova2006beyond}, to have a general way of comparing our test results, we altered the positive examples between the majority and minority classes when computing $\fscore$ and $\gscore$. Thus we show $\fscore$ and $\gscore$ for the majority and the minority classes separately.

\begin{table*}[]
\begin{center}
\scalebox{1.0}{
\begin{tabular}{lc!{\VRule[1.5pt]}c|c|c|c!{\VRule[1.5pt]}c|c|c|c!{\VRule[1.5pt]}}
\Cline{1.5pt}{3-10}
 & \textbf{} & \multicolumn{4}{c!{\VRule[1.5pt]}}{\textbf{Minority}} & \multicolumn{4}{c!{\VRule[1.5pt]}}{\textbf{Majority}} \\ \Cline{1.5pt}{2-10} 
\multicolumn{1}{l!{\VRule[1.5pt]}}{} & \textbf{AUC} & \textbf{Precision} & \textbf{Recall} & \textbf{Fscore} & \textbf{Gscore} & \textbf{Precision} & \textbf{Recall} & \textbf{Fscore} & \textbf{Gscore} \\
\specialrule{1.5pt}{0pt}{0pt}
\multicolumn{1}{!{\VRule[1.5pt]}l!{\VRule[1.5pt]}}{\textbf{b-kde}} & &  \multicolumn{4}{c!{\VRule[1.5pt]}}{} & \multicolumn{4}{c!{\VRule[1.5pt]}}{}  \\               
\specialrule{1.5pt}{0pt}{0pt}                     
\multicolumn{1}{!{\VRule[1.5pt]}l!{\VRule[1.5pt]}}{\textbf{Original}} & 0.797 & 0.139 & 0.033 & 0.054 & 0.068 & 0.830 & \textbf{0.995} & \textbf{0.905} & \textbf{0.909}  \\ \hline
\multicolumn{1}{!{\VRule[1.5pt]}l!{\VRule[1.5pt]}}{\textbf{Dup}} & 0.733 & 0.385 & 0.454 & 0.417 & 0.418 & 0.869 & 0.742 & 0.801 & 0.803  \\ \hline
\multicolumn{1}{!{\VRule[1.5pt]}l!{\VRule[1.5pt]}}{\textbf{SMOTE}} & 0.807 & 0.488 & 0.705 & 0.577 & \textbf{0.587} & 0.833 & 0.644 & 0.726 & 0.733  \\ \hline
\multicolumn{1}{!{\VRule[1.5pt]}l!{\VRule[1.5pt]}}{\textbf{B-SMOTE}} & 0.774 & 0.258 & 0.456 & 0.330 & 0.343 & 0.846 & 0.671 & 0.748 & 0.754  \\ \hline
\multicolumn{1}{!{\VRule[1.5pt]}l!{\VRule[1.5pt]}}{\textbf{MWMOTE}} & 0.794 & 0.396 & \textbf{0.754} & 0.520 & 0.547 & 0.836 & 0.557 & 0.669 & 0.682  \\ \hline
\multicolumn{1}{!{\VRule[1.5pt]}l!{\VRule[1.5pt]}}{\textbf{ADASYN}} & 0.802 & 0.395 & 0.632 & 0.487 & 0.500 & 0.817 & 0.598 & 0.691 & 0.699  \\ \hline
\multicolumn{1}{!{\VRule[1.5pt]}l!{\VRule[1.5pt]}}{\textbf{RAMOboost}} & 0.748 & 0.358 & 0.343 & 0.350 & 0.350 & 0.860 & 0.822 & 0.841 & 0.841  \\ \hline
\multicolumn{1}{!{\VRule[1.5pt]}l!{\VRule[1.5pt]}}{\textbf{CGMOS}} & \textbf{0.842} & \textbf{0.536} & 0.517 & \textbf{0.526} & 0.526 & \textbf{0.908} & 0.815 & 0.859 & 0.860  \\ 
\specialrule{1.5pt}{0pt}{0pt}                     
\multicolumn{1}{!{\VRule[1.5pt]}l!{\VRule[1.5pt]}}{\textbf{knn}} & & \multicolumn{4}{c|}{} & \multicolumn{4}{c!{\VRule[1.5pt]}}{}  \\               
\specialrule{1.5pt}{0pt}{0pt}                     
\multicolumn{1}{!{\VRule[1.5pt]}l!{\VRule[1.5pt]}}{\textbf{Original}} & 0.821 & \textbf{0.701} & 0.521 & 0.598 & 0.604 & 0.902 & \textbf{0.942} & \textbf{0.922} & \textbf{0.9217}  \\ \hline
\multicolumn{1}{!{\VRule[1.5pt]}l!{\VRule[1.5pt]}}{\textbf{Dup}} & 0.810 & 0.519 & 0.732 & 0.607 & 0.616 & 0.921 & 0.818 & 0.867 & 0.868  \\ \hline
\multicolumn{1}{!{\VRule[1.5pt]}l!{\VRule[1.5pt]}}{\textbf{SMOTE}} & 0.827 & 0.506 & \textbf{0.804} & 0.621 & 0.638 & 0.925 & 0.805 & 0.861 & 0.863  \\ \hline
\multicolumn{1}{!{\VRule[1.5pt]}l!{\VRule[1.5pt]}}{\textbf{B-SMOTE}} & 0.811 & 0.494 & 0.736 & 0.591 & 0.603 & 0.927 & 0.790 & 0.853 & 0.856  \\ \hline
\multicolumn{1}{!{\VRule[1.5pt]}l!{\VRule[1.5pt]}}{\textbf{MWMOTE}} & 0.832 & 0.504 & 0.792 & 0.616 & 0.632 & \textbf{0.928} & 0.794 & 0.856 & 0.858  \\ \hline
\multicolumn{1}{!{\VRule[1.5pt]}l!{\VRule[1.5pt]}}{\textbf{ADASYN}} & 0.825 & 0.495 & 0.786 & 0.607 & 0.623 & \textbf{0.929} & 0.786 & 0.851 & 0.854  \\ \hline
\multicolumn{1}{!{\VRule[1.5pt]}l!{\VRule[1.5pt]}}{\textbf{RAMOboost}} & 0.827 & 0.540 & 0.684 & 0.604 & 0.608 & 0.918 & 0.847 & 0.881 & 0.881  \\ \hline
\multicolumn{1}{!{\VRule[1.5pt]}l!{\VRule[1.5pt]}}{\textbf{CGMOS}} & \textbf{0.840} & 0.544 & 0.766 & \textbf{0.636} & \textbf{0.646} & 0.925 & 0.842 & 0.882 & 0.883  \\ 
\specialrule{1.5pt}{0pt}{0pt}                     
\multicolumn{1}{!{\VRule[1.5pt]}l!{\VRule[1.5pt]}}{\textbf{svm}} & & \multicolumn{4}{c|}{} & \multicolumn{4}{c!{\VRule[1.5pt]}}{}  \\               
\specialrule{1.5pt}{0pt}{0pt}                     
\multicolumn{1}{!{\VRule[1.5pt]}l!{\VRule[1.5pt]}}{\textbf{Original}} & 0.792 & \textbf{0.632} & 0.587 & 0.609 & 0.609 & 0.882 & 0.935 & 0.908 & 0.908  \\ \hline
\multicolumn{1}{!{\VRule[1.5pt]}l!{\VRule[1.5pt]}}{\textbf{Dup}} & 0.815 & 0.543 & 0.436 & 0.484 & 0.487 & \textbf{0.981} & 0.861 & 0.917 & 0.919  \\ \hline
\multicolumn{1}{!{\VRule[1.5pt]}l!{\VRule[1.5pt]}}{\textbf{SMOTE}} & 0.844 & 0.579 & 0.726 & 0.644 & 0.648 & 0.879 & 0.844 & 0.861 & 0.861  \\ \hline
\multicolumn{1}{!{\VRule[1.5pt]}l!{\VRule[1.5pt]}}{\textbf{B-SMOTE}} & 0.832 & 0.475 & 0.729 & 0.575 & 0.588 & 0.893 & \textbf{0.959} & \textbf{0.924} & \textbf{0.925}  \\ \hline
\multicolumn{1}{!{\VRule[1.5pt]}l!{\VRule[1.5pt]}}{\textbf{MWMOTE}} & 0.830 & 0.547 & 0.647 & 0.593 & 0.595 & 0.880 & 0.884 & 0.882 & 0.882  \\ \hline
\multicolumn{1}{!{\VRule[1.5pt]}l!{\VRule[1.5pt]}}{\textbf{ADASYN}} & 0.827 & 0.536 & 0.654 & 0.589 & 0.592 & 0.880 & 0.755 & 0.813 & 0.815  \\ \hline
\multicolumn{1}{!{\VRule[1.5pt]}l!{\VRule[1.5pt]}}{\textbf{RAMOboost}} & 0.842 & 0.556 & 0.673 & 0.609 & 0.611 & 0.968 & 0.852 & 0.906 & 0.908  \\ \hline
\multicolumn{1}{!{\VRule[1.5pt]}l!{\VRule[1.5pt]}}{\textbf{CGMOS}} & \textbf{0.864} & 0.555 & \textbf{0.788} & \textbf{0.651} & \textbf{0.661} & 0.943 & 0.830 & 0.883 & 0.885  \\ 
\specialrule{1.5pt}{0pt}{0pt}                     
\multicolumn{1}{!{\VRule[1.5pt]}l!{\VRule[1.5pt]}}{\textbf{nn}} & & \multicolumn{4}{c|}{} & \multicolumn{4}{c!{\VRule[1.5pt]}}{}  \\               
\specialrule{1.5pt}{0pt}{0pt}                     
\multicolumn{1}{!{\VRule[1.5pt]}l!{\VRule[1.5pt]}}{\textbf{Original}} & 0.801 & \textbf{0.632} & 0.412 & 0.499 & 0.510 & 0.892 & \textbf{0.962} & \textbf{0.925} & \textbf{0.9258}  \\ \hline
\multicolumn{1}{!{\VRule[1.5pt]}l!{\VRule[1.5pt]}}{\textbf{Dup}} & 0.843 & 0.543 & 0.777 & 0.639 & 0.650 & 0.926 & 0.819 & 0.869 & 0.871  \\ \hline
\multicolumn{1}{!{\VRule[1.5pt]}l!{\VRule[1.5pt]}}{\textbf{SMOTE}} & 0.840 & 0.555 & 0.750 & 0.638 & 0.645 & 0.921 & 0.820 & 0.868 & 0.869  \\ \hline
\multicolumn{1}{!{\VRule[1.5pt]}l!{\VRule[1.5pt]}}{\textbf{B-SMOTE}} & 0.841 & 0.475 & 0.779 & 0.590 & 0.608 & 0.924 & 0.802 & 0.859 & 0.861  \\ \hline
\multicolumn{1}{!{\VRule[1.5pt]}l!{\VRule[1.5pt]}}{\textbf{MWMOTE}} & 0.841 & 0.547 & 0.778 & 0.642 & 0.652 & 0.927 & 0.812 & 0.866 & 0.867  \\ \hline
\multicolumn{1}{!{\VRule[1.5pt]}l!{\VRule[1.5pt]}}{\textbf{ADASYN}} & 0.842 & 0.536 & \textbf{0.786} & 0.637 & 0.649 & 0.929 & 0.803 & 0.861 & 0.863  \\ \hline
\multicolumn{1}{!{\VRule[1.5pt]}l!{\VRule[1.5pt]}}{\textbf{RAMOboost}} & 0.841 & 0.556 & 0.743 & 0.636 & 0.643 & 0.919 & 0.830 & 0.872 & 0.873  \\ \hline
\multicolumn{1}{!{\VRule[1.5pt]}l!{\VRule[1.5pt]}}{\textbf{CGMOS}} & \textbf{0.865} & 0.579 & 0.750 & \textbf{0.653} & \textbf{0.659} & \textbf{0.933} & 0.845 & 0.887 & 0.888  \\ 
\specialrule{1.5pt}{0pt}{0pt}                     
\multicolumn{1}{!{\VRule[1.5pt]}l!{\VRule[1.5pt]}}{\textbf{rf}} & & \multicolumn{4}{c|}{} & \multicolumn{4}{c!{\VRule[1.5pt]}}{}  \\               
\specialrule{1.5pt}{0pt}{0pt}                     
\multicolumn{1}{!{\VRule[1.5pt]}l!{\VRule[1.5pt]}}{\textbf{Original}} & 0.872 & \textbf{0.699} & 0.534 & 0.606 & 0.611 & 0.909 & \textbf{0.956} & \textbf{0.932} & \textbf{0.932}  \\ \hline
\multicolumn{1}{!{\VRule[1.5pt]}l!{\VRule[1.5pt]}}{\textbf{Dup}} & 0.873 & 0.682 & 0.641 & 0.661 & 0.661 & 0.917 & 0.924 & 0.921 & 0.921  \\ \hline
\multicolumn{1}{!{\VRule[1.5pt]}l!{\VRule[1.5pt]}}{\textbf{SMOTE}} & 0.875 & 0.667 & 0.655 & 0.661 & 0.661 & 0.920 & 0.917 & 0.918 & 0.918  \\ \hline
\multicolumn{1}{!{\VRule[1.5pt]}l!{\VRule[1.5pt]}}{\textbf{B-SMOTE}} & 0.867 & 0.653 & 0.637 & 0.645 & 0.645 & 0.920 & 0.906 & 0.913 & 0.913  \\ \hline
\multicolumn{1}{!{\VRule[1.5pt]}l!{\VRule[1.5pt]}}{\textbf{MWMOTE}} & 0.878 & 0.658 & 0.651 & 0.655 & 0.655 & 0.920 & 0.922 & 0.921 & 0.921  \\ \hline
\multicolumn{1}{!{\VRule[1.5pt]}l!{\VRule[1.5pt]}}{\textbf{ADASYN}} & 0.876 & 0.663 & 0.669 & 0.666 & 0.666 & 0.919 & 0.915 & 0.917 & 0.917  \\ \hline
\multicolumn{1}{!{\VRule[1.5pt]}l!{\VRule[1.5pt]}}{\textbf{RAMOboost}} & 0.874 & 0.686 & 0.618 & 0.650 & 0.651 & 0.915 & 0.933 & 0.924 & 0.924  \\ \hline
\multicolumn{1}{!{\VRule[1.5pt]}l!{\VRule[1.5pt]}}{\textbf{CGMOS}} & \textbf{0.884} & 0.685 & \textbf{0.678} & \textbf{0.681} & \textbf{0.681} & \textbf{0.923} & 0.926 & 0.924 & 0.924  \\ 
\specialrule{1.5pt}{0pt}{0pt}                     
\multicolumn{1}{!{\VRule[1.5pt]}l!{\VRule[1.5pt]}}{\textbf{Adaboost.M1}} & & \multicolumn{4}{c|}{} & \multicolumn{4}{c!{\VRule[1.5pt]}}{}  \\               
\specialrule{1.5pt}{0pt}{0pt}                     
\multicolumn{1}{!{\VRule[1.5pt]}l!{\VRule[1.5pt]}}{\textbf{Original}} & 0.868 & \textbf{0.699} & 0.572 & 0.629 & 0.632 & 0.906 & \textbf{0.944} & \textbf{0.925} & \textbf{0.9247}  \\ \hline
\multicolumn{1}{!{\VRule[1.5pt]}l!{\VRule[1.5pt]}}{\textbf{Dup}} & 0.865 & 0.622 & 0.708 & 0.662 & 0.664 & 0.922 & 0.873 & 0.897 & 0.897  \\ \hline
\multicolumn{1}{!{\VRule[1.5pt]}l!{\VRule[1.5pt]}}{\textbf{SMOTE}} & 0.867 & 0.608 & 0.714 & 0.657 & 0.659 & 0.923 & 0.880 & 0.901 & 0.901  \\ \hline
\multicolumn{1}{!{\VRule[1.5pt]}l!{\VRule[1.5pt]}}{\textbf{B-SMOTE}} & 0.864 & 0.581 & 0.724 & 0.644 & 0.648 & \textbf{0.927} & 0.861 & 0.893 & 0.893  \\ \hline
\multicolumn{1}{!{\VRule[1.5pt]}l!{\VRule[1.5pt]}}{\textbf{MWMOTE}} & 0.868 & 0.600 & 0.708 & 0.650 & 0.652 & 0.922 & 0.880 & 0.901 & 0.901  \\ \hline
\multicolumn{1}{!{\VRule[1.5pt]}l!{\VRule[1.5pt]}}{\textbf{ADASYN}} & 0.867 & 0.599 & 0.726 & 0.657 & 0.660 & 0.925 & 0.873 & 0.898 & 0.899  \\ \hline
\multicolumn{1}{!{\VRule[1.5pt]}l!{\VRule[1.5pt]}}{\textbf{RAMOboost}} & 0.865 & 0.631 & 0.699 & 0.663 & 0.664 & 0.922 & 0.882 & 0.901 & 0.902  \\ \hline
\multicolumn{1}{!{\VRule[1.5pt]}l!{\VRule[1.5pt]}}{\textbf{CGMOS}} & \textbf{0.871} & 0.619 & \textbf{0.728} & \textbf{0.670} & \textbf{0.672} & 0.925 & 0.882 & 0.903 & 0.903  \\ 
\specialrule{1.5pt}{0pt}{0pt}
\end{tabular}
}
\end{center}
\caption{A summary of AUC, $Precision$, $Recall$, $\fscore$ and $\gscore$ of all competitors for the majority and minority classes produced by 6 classifiers on the artificial datasets.}
\label{tab: results_real_data}
\end{table*}

\begin{table*}[t]
\begin{center}
\scalebox{1.0}{
\centering
\begin{tabular}{!{\VRule[1.5pt]}l!{\VRule[1.5pt]}c|c|c|c|c|c|c|c!{\VRule[1.5pt]}}
\specialrule{1.5pt}{0pt}{0pt} 
 & \textbf{CGMOS} & \textbf{Original} & \textbf{Dup} & \textbf{SMOTE} & \textbf{B-SMOTE} & \textbf{MWMOTE} & \textbf{ADASYN} & \textbf{RAMOboost}\\
 \specialrule{1.5pt}{0pt}{0pt} 
\textbf{BankMarket} & \textbf{0.728} & 0.661 & 0.708 & 0.718 & 0.710 & 0.721 & 0.710 & 0.723 \\ \hline
\textbf{BloodService} & \textbf{0.733} & 0.653 & 0.648 & 0.649 & 0.651 & 0.720 & 0.714 & 0.728 \\ \hline
\textbf{BreastCancer} & 0.992 & 0.992 & \textbf{0.993} & 0.992 & 0.989 & 0.991 & 0.991 & 0.992 \\ \hline
\textbf{BreastTissue} & \textbf{0.984} & 0.899 & 0.946 & 0.932 & 0.917 & 0.937 & 0.908 & 0.943 \\ \hline
\textbf{CarEvaluation} & \textbf{0.997} & 0.995 & 0.845 & \textbf{0.997} & 0.994 & 0.996 & \textbf{0.997} & 0.995 \\ \hline
\textbf{Card'graphy} & \textbf{0.977} & 0.976 & 0.939 & 0.962 & 0.956 & 0.925 & 0.957 & 0.960 \\ \hline
\textbf{CharacterTraj} & 0.985 & 0.962 & 0.717 & 0.985 & 0.978 & 0.981 & \textbf{0.988} & 0.909 \\ \hline
\textbf{Chess} & \textbf{0.977} & 0.974 & 0.959 & 0.973 & \textbf{0.977} & 0.974 & 0.975 & 0.959 \\ \hline
\textbf{ClimateSim} & \textbf{0.908} & \textbf{0.908} & 0.861 & 0.902 & 0.863 & 0.901 & 0.901 & 0.882 \\ \hline
\textbf{Contraceptive} & \textbf{0.724} & 0.705 & 0.699 & 0.712 & 0.702 & 0.705 & 0.702 & 0.705 \\ \hline
\textbf{Fertility} & \textbf{0.673} & 0.615 & 0.594 & 0.634 & 0.592 & 0.604 & 0.639 & 0.638 \\ \hline
\textbf{Haberman} & \textbf{0.651} & 0.623 & 0.577 & 0.600 & 0.593 & 0.594 & 0.587 & 0.586 \\ \hline
\textbf{ILPD} & 0.707 & 0.687 & 0.693 & \textbf{0.715} & 0.703 & 0.702 & 0.693 & 0.703 \\ \hline
\textbf{ImgSeg} & \textbf{0.999} & 0.998 & \textbf{0.999} & 0.997 & 0.998 & 0.998 & 0.997 & 0.998 \\ \hline
\textbf{Leaf} & \textbf{0.908} & 0.880 & 0.782 & 0.852 & 0.775 & 0.836 & 0.839 & 0.821 \\ \hline
\textbf{Libras} & \textbf{0.945} & 0.922 & 0.859 & 0.929 & 0.886 & 0.936 & 0.923 & 0.883 \\ \hline
\textbf{MultipleFs} & \textbf{0.998} & \textbf{0.998} & 0.997 & \textbf{0.998} & 0.997 & 0.997 & 0.996 & 0.997 \\ \hline
\textbf{Parkinson} & 0.841 & 0.676 & 0.692 & 0.834 & 0.791 & 0.837 & \textbf{0.842} & 0.760 \\ \hline
\textbf{PlanRelax} & 0.472 & 0.457 & 0.494 & 0.469 & 0.445 & 0.467 & \textbf{0.488} & 0.464 \\ \hline
\textbf{QSAR} & \textbf{0.901} & 0.886 & 0.879 & 0.895 & 0.863 & 0.886 & 0.886 & 0.882 \\ \hline
\textbf{SPECT} & \textbf{0.820} & 0.772 & 0.803 & 0.808 & 0.811 & 0.752 & 0.801 & 0.799 \\ \hline
\textbf{SPECTF} & 0.819 & 0.819 & 0.800 & 0.805 & 0.816 & 0.812 & \textbf{0.825} & 0.795 \\ \hline
\textbf{SeismicBumps} & \textbf{0.743} & 0.735 & 0.712 & 0.727 & 0.740 & 0.732 & 0.715 & 0.691 \\ \hline
\textbf{Statlog} & \textbf{0.998} & 0.992 & 0.996 & \textbf{0.998} & 0.990 & 0.996 & 0.976 & 0.996 \\ \hline
\textbf{PlatesFaults} & \textbf{0.956} & 0.928 & 0.844 & 0.954 & 0.920 & 0.943 & \textbf{0.956} & 0.881 \\ \hline
\textbf{TAEvaluation} & \textbf{0.748} & 0.682 & 0.644 & 0.703 & 0.671 & 0.707 & 0.665 & 0.657 \\ \hline
\textbf{UserKnowledge} & \textbf{0.958} & 0.837 & 0.919 & 0.953 & 0.947 & 0.951 & 0.950 & 0.888 \\ \hline
\textbf{Vertebral} & \textbf{0.890} & 0.839 & 0.869 & 0.855 & 0.829 & 0.860 & 0.794 & 0.872 \\ \hline
\textbf{Customers} & \textbf{0.952} & 0.930 & 0.943 & 0.946 & 0.884 & 0.902 & 0.946 & \textbf{0.952} \\ \hline
\textbf{Yeast} & \textbf{0.925} & 0.792 & 0.844 & 0.907 & 0.898 & 0.900 & 0.906 & 0.851 \\
\specialrule{1.5pt}{0pt}{0pt} 
\textbf{Average} & \textbf{0.864} & 0.827 & 0.808 & 0.844 & 0.830 & 0.842 & 0.842 & 0.830 \\
\specialrule{1.5pt}{0pt}{0pt} 
\end{tabular}
}
\end{center}
\caption{A summary of AUC of 8 oversampling algorithms over all 30 datasets used in our evaluation. The AUC is averaged over all 6 base classifiers used in the evaluation. It could be seen from above table that CGMOS achieves best AUC measures for 24 datasets out of 30. By average, the AUC of CGMOS is at least 2 percent higher than all other competitors.}
\label{tab: results_real_data_per_file}
\end{table*} 

Although, both $\fscore$ and $\gscore$ are great evaluation metrics, they are still less effective in some situations. So we also employ the ROC graphs \cite{fawcett2004roc}\cite{fawcett2006introduction}\cite{mohri2005confidence} in the evaluation. ROC graph is a two-dimensional graph, while $FP$ $rate$ and $TP$ $rate$ are its X axis and Y axis respectively.

An ROC graph basically manifest its usefulness by showing relative trade-off between benefit (true positive) and cost (false positive). One attractive property make ROC graph a good metric in imbalanced learning lies in the facts that ROC curve is insensitive to changes in class distribution. Because of this property, it is easier to see the performances of models trained by dataset oversampled by different algorithms. The goal in ROC space is to let curves be as close to upper-left-hand corner as possible, in which case the ratio between benefit and cost is maximized. To compare all test results in a more straightforward way, we also compute area under an ROC curve (AUC) which reduce the ROC performance to a single scalar value representing expected performance of the ROC curve.

\subsection{Results}
This section presents the performance of CGMOS and all the other methods on 30 real-world datasets. The same experiment procedure as the one in the experiments of the artificial dataset was conducted. All results are averged from 10 rounds of 10-folds cross-validations. A summary of the experiment results is shown in Table \ref{tab: results_real_data} and ROC graphs are shown in Figure \ref{fig: roc}.

\begin{table}[ht]
\begin{center}
\scalebox{0.75}{
\begin{tabular}{!{\VRule[1.5pt]}l!{\VRule[1.5pt]}c|c|c|c|c|c!{\VRule[1.5pt]}}
\specialrule{1.5pt}{0pt}{0pt} 
 & \makebox[3em]{\textbf{Knn}} & \makebox[3em]{\textbf{Rf}}& \makebox[3em]{\textbf{B-kde}} & \makebox[3em]{\textbf{Nn}} & \makebox[3em]{\textbf{Svm}} & \makebox[3em]{\textbf{Boost}}\\
\specialrule{1.5pt}{0pt}{0pt} 
\textbf{Original} & 5e-5 & 1e-4 & 0.004 & 1e-4 & 0.026 & 0.04\\\hline
\textbf{Dup} & 2e-6 & 5e-5 & 3e-6 & 0.03 & 0.049 & 0.004\\\hline
\textbf{SMOTE} & 0.003 & 2e-4 & 6e-6 & 0.018 & 0.006 & 0.046\\\hline
\textbf{B-SMOTE} & 4e-6 & 7e-6 & 2e-5 & 5e-4 & 0.047 & 5e-4\\\hline
\textbf{MWMOTE} & 0.046 & 4e-5 & 1e-5 & 0.003 & 0.005 & 0.007\\\hline
\textbf{ADASYN} & 8e-6 & 7e-5 & 9e-5 & 0.005 & 1e-4 & 0.003\\\hline
\textbf{RAMOboost} & 2e-6 & 5e-5 & 3e-6 & 0.001 & 0.045 & 0.035\\
\specialrule{1.5pt}{0pt}{0pt} 
\end{tabular}
}
\end{center}
\caption{A summary of $p$-values of statistical significant tests of classification results using CGMOS against each of all the other competitors. }
\label{tab: signrank}
\end{table}

Considering the classification results of the minority class, it can be observed that the proposed approach outperforms most of the compared methods under all classification algorithms in terms of $\fscore$ and $\gscore$. For $\fscore$ and $\gscore$ of the majority class, the proposed approach in most cases is only second to the original data without oversampling. This is because the original dataset is imbalanced and it favors the majority class more than the minority class during classification. Overall, CGMOS achieves the best AUC over all tests. This is because the proposed approach takes into account both of the majority and minority classes and increases the certainties of the two classes while oversampling.

The same conclusion can be made from the ROC curves shown in Fig. \ref{fig: roc}. It could be seen from the ROC curves that the proposed approach has the highest values almost everywhere. The proposed approach achieves the best result when random forest is used as the classifier. For b-kde as the classifier, the proposed approach gets the largest improvement
since the design of the proposed approach uses b-kde for certainty computations.

%

To get a closer view of the performances of all compared methods on each dataset, we show the AUC results of CGMOS and all other compared methods for each dataset in Table \ref{tab: results_real_data_per_file}. The table shows that by average the AUC of CGMOS is 2 percent higher than SMOTE whose AUC is 2nd highest. 

Previous studies show that it is not necessary for a learning procedure to obtain best classification results when a dataset is perfectly balanced\cite{batista2004study}\cite{weiss2003learning}. How much to oversample is usually empirically determined \cite{chawla2004editorial}. To evaluate this aspect we performed another experiment in which we synthesized increasing number of minority samples and investigated how different amounts of new samples impact classification results.

Let $\delta$ denote the difference of data samples between the majority and the minority class. We performed multiple experiments where in each round we synthesized $k\delta$ new samples of the  minority class where $k$ gradually increased from $0.5$ to $5$. The classification results are shown in Figure \ref{fig: increasingnum}. As can be observed in the results, CGMOS achieves the best results in all cases. Also, observe that when increasing the number of data samples added, the results of CGMOS are much more robust compared with other approaches. Note that the results of some methods such as Dup(b-kde), B-SMOTE(knn) and B-SMOTE(Adaboost.M1) are even lower than the results at the starting point where datasets are not oversampled. This highlights the advantage of CGMOS when handling oversampling on boundary samples. 

\subsection{Statistical Significance Analysis}
We evaluate the statistical significance of the classification results of all competitors. Statistical significance plays a critical role in determining whether a null hypothesis should be rejected or retained, where the term null hypothesis refers to a general statement that sample observations result purely from chance. For a null hypothesis to be rejected as false, the result has to be identified as being statistically significant.

To determine whether to reject a null hypothesis, a $p$-value has to be calculated, which is the probability of observing an effect given that the null hypothesis is true \cite{JLD:11}. The null hypothesis is rejected if $p$-value is less than the significance level. The significance level is the probability of rejecting the null hypothesis given that it is true. The lower the significance level the more confident we can be in replicating the results and usually the significance level is set at $5\%$. Then a sample observation is determined to be statistically significant if $p$-value is less than $5\%$, which is formally written as $p<0.05$ \cite{SM:06}.

We follow the same protocols used in \cite{demvsar2006statistical}\cite{chen2010ramoboost}\cite{barua2014mwmote} and choose to use Wilcoxon signed-ranks test in this paper. Wilcoxon signed-ranks test is a nonparametric statistical procedure for comparing two samples that are paired, or related \cite{GC09}. Different from $t$-test \cite{BF:08}\cite{zimmerman1997teacher}\cite{demvsar2006statistical} whose null hypothesis is that the mean difference between pairs is zero, the null hypothesis of Wilcoxon signed-ranks test is that the median difference between pairs of observations is zero.

The test results are shown in Table \ref{tab: signrank}. It could be seen from the table that the $p$-value of all tests are smaller than 0.05 and pass the test. 

\section{Conclusion}
In this paper, we address the imbalanced binary classification problem by proposing a novel minority oversampling strategy. Different from existing approaches, CGMOS does not randomly synthesize new data along decision boundaries. Instead, CGMOS computes the Bayes classification certainties for both the majority and minority classes and then synthesize new samples based on improvement of the certainties for samples in both classes. We prove that CGMOS can achieve better classification results compared with SMOTE. In addition, experimental results show that CGMOS outperforms known oversampling techniques using various metrics.

{\small
\bibliographystyle{IEEEtran}
\bibliography{mybib}
}

\end{document}